# A Review of Visual Trackers and Analysis of its Application to Mobile Robot

Shaoze You[1,2], Hua Zhu[1,2,*], Menggang Li[1,2], Yutan Li[1,2],

[1] School of Mechanical and Electrical Engineering, China University of Mining and Technology, Xuzhou 221116, China; youshaoze@cumt.edu.cn
[2] Jiangsu Collaborative Innovation Center of Intelligent Mining Equipment, China University of Mining and Technology, Xuzhou 221008, China;
* Correspondence: zhuhua83591917@163.com; Tel.: +86-0516-83591917

**Abstract:** Computer vision has received a significant attention in recent year, which is one of the important parts for robots to obtain information about the external environment. Visual trackers can provide the necessary physical and environmental parameters for the mobile robot, and their performance is related to the actual application of the robot. This study provides a comprehensive survey on visual trackers. Following a brief introduction, we first analyzed the basic framework and difficulties of visual trackers. Then the structure of generative and discriminative methods is introduced, and summarized the feature descriptors, modeling methods, and learning methods which be used in tracker. Later we reviewed and evaluated the state-of-the-art progress on discriminative trackers from three directions: correlation filter, deep learning and convolutional features. Finally, we analyzed the research direction of visual tracker used in mobile robot, as well as outlined the future trends for visual tracker on mobile robot.

## 1. Introduction

The eye is an important organ for human to get information from the outside world. According to statistics [1], nearly 80% of the environmental information (color, brightness, shape, movement, depth, etc.) comes from vision. Computer vision (CV) gives computers the ability to "see the world" like humans. It uses cameras to mimic the function of the human eye, so as to realize the functions of extraction, recognition and tracking of the object. Visual tracking is one of the most challenging problems in computer vision, it can provide robot with tracking, location and recognition of the specified target, and the parameters of the target or environment can be provided to the controller for subsequent use. It enjoys wide applications in the field of machine intelligence, including in mobile robotics, autonomous driving, human computer interaction, automated surveillance and Eye-tracking technology etc.

### 1.1  Tracking algorithm and visual tracker

The traditional tracking algorithm is different from the visual tracker in CV. The former is more suitable as tracking strategy. This kind of algorithm can predict the moving state of the target in the next frame by putting forward mathematical formula to model the change of the state space of the target in time domain. The latter is the integration of detection algorithm, tracking strategy, update strategy, online classifier, re-detector and other branch algorithms in CV, which has a more complex system structure. In this paper, the related work of the latter was introduced and analyzed emphatically.

## 1.2 Aim and outline

As one of the research hotspots in the field of computer vision, to evaluate the synthetic performance of visual trackers, starting with PETS [2] and VIVID [3], many researchers have provided evaluation datasets, and many people have proposed tracking training set [4-6] (shown in Table 1). From Wu's evaluation benchmark [7, 8] to the VOT [9-14] visual competition, the performance of state-of-the-art visual trackers have been ranked, and some of the tracker have been open source. We combined the data of the evaluation database as a reference, firstly we introduced the difficulties and basic framework of visual tracking in Section 2. In Section 3, state-of-the-art trackers based on tracking-by-detection were summarized. Section 4 was dedicated to analyze the characteristics of tracker required in the field of mobile robot. At last, conclusion and future directions could be found in Section 5.

Table 1. Datasets proposed in recent years

| Dataset | Year | Videos | Duration | Frame rate |
|---|---|---|---|---|
| PETS [2] | 2004 | 28 | ST | 30FPS |
| VIVID [3] | 2005 | 9 | LT | 30FPS |
| OTB-50 [7] | 2013 | 50 | ST | 30FPS |
| PTB [15] | 2013 | 100 | ST | 30FPS |
| ALOV++ [16] | 2013 | 314 | ST | 30FPS |
| VOT [12] | 2014-2018 | 25, 60, $60^1$, $60^2$, $60^3$ | ST | 30FPS |
| TC-128 [17] | 2015 | 129 | ST | 30FPS |
| OTB-100 [8] | 2015 | 100 | ST | 30FPS |
| NUS-PRO [18] | 2015 | 365 | ST | 30FPS |
| VID [4] | 2015 | 4,417 Tr | ST & LT | - |
| UAV-123 [19] | 2016 | 123S + 20L | ST & LT | 30FPS |
| NfS [20] | 2017 | 100 | ST | 240FPS |
| DTB-70 [21] | 2017 | 70 | ST | 30FPS |
| AMP [22] | 2017 | 100 | ST | 30FPS |
| TLP [23] | 2017 | 50 | LT | 24/30FPS |
| YTBB [5] | 2017 | 380,000 Tr | ST & LT | - |
| VOT-LT [24] | 2018 | 35 | LT | 30FPS |
| TrackingNet [6] | 2018 | 30,132 Tr + 511 Te | ST & LT | - |
| LTWB [25] | 2018 | 366 | LT | - |

Note: The VOT competition has changed every year, recalibrating in 2016, replacing 10 easy sequences with 10 difficult ones in 2017 and adding 35 Long-Term tracking sequences VOT-LT in 2018. ST: Short-Term, LT: Long-Term, Tr: Training set, Te: Test set.

## 2. Visual tracker architecture and classification

### 2.1 Basic Framework and the problems in tracking system

Visual tracking has developed significantly over the past few decades [26-32], and the process of visual tracking has been clear since it was first put forward to now. For an input video or image sequence, firstly, the state of the current frame of the target is taken as the initial state of tracking (initialization model parameter), and then the key points are extracted and modeled. Then the target model is applied to the subsequent frames, and the current state of the target is estimated by the tracking strategy (filtering method, optical flow method, etc.). Further, the target model is updated by the current state. Finally, tracking the target model in the next frame. The basic flowchart of visual tracking is shown in Figure 1.

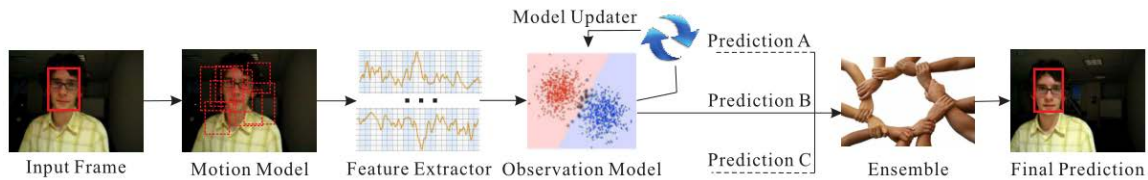

**Fig. 1. The Framework flow of a Visual tracking system.** Naiyan Wang et al. [33] divided the traditional visual tracking algorithm framework in detail. They decomposed the visual tracking into five parts: **Motion Model, Feature Extractor, Observation Model, Model Updater, Ensemble Post-processor**. Then the experimental results were shown that feature extraction is far more important than observational model in visual tracking.

In the above tracking framework, feature extractor is the process of describing the target. On the basis of the extracted target feature, the object description model is constructed. Tracker can be divided into two categories according to the way of target feature extraction and observation model (online learning method): **Generative** method and **Discriminative** method. The method used to predict the trajectory of a target in the observation model is the tracking strategy, such as Kalman filter [34], extended Kalman filter [35], particle filter [36], L-K optical flow algorithm [37], Markov chain Monte Carlo algorithm [38], Normalized Cross Correlation [39], Mean-Shift [28, 40] and Cam-shift [41]. In the process of visual tracking, the state of the target and its surrounding environment are constantly changing (Figure 2), which not only makes it difficult to extract features and build models, but also requires trackers have more robustness and higher accuracy. Based on this, real-time tracking is also possible.

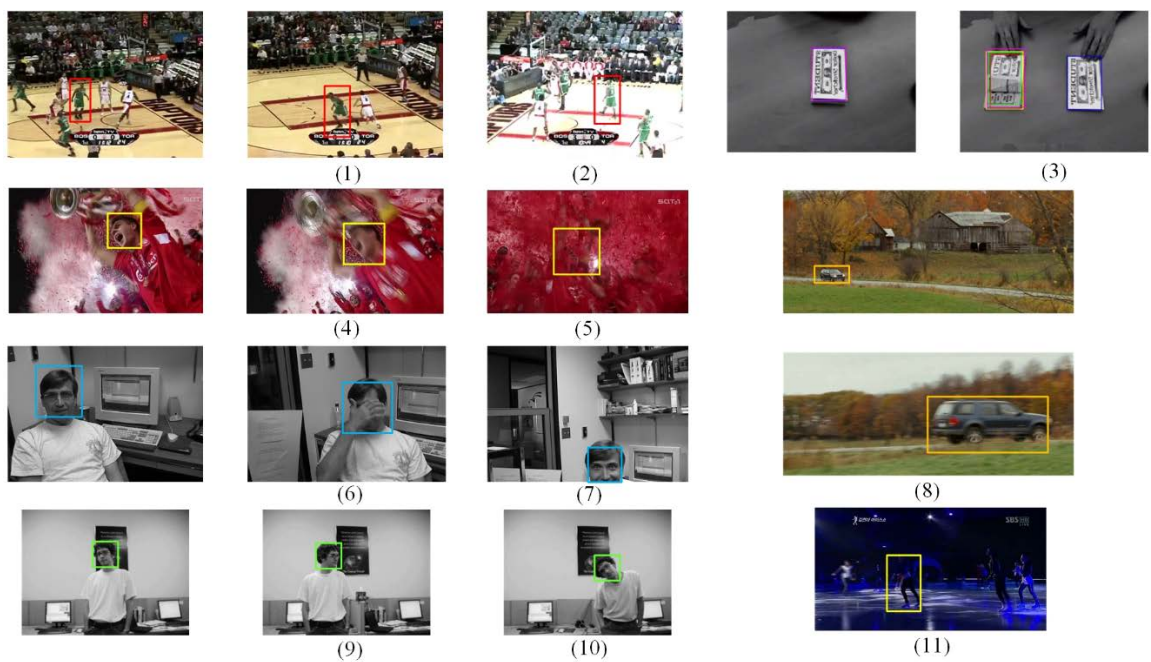

**Fig.2. Challenges and difficulties in Visual tracking.** There are generally recognized difficulties in tracking: (1) Appearance deformation; (2) Illumination change; (3) Appearance similarity; (4) Motion blur; (5) Background clutter; (6) Occlusion; (7) Out of view; (8) Scale change; (9) Out of plane rotation; (10) In plane rotation; (11) Background similar.

## 2.2 Generative Method

In the process of learning, generative method is to obtain conditional probability distribution $P(Y$

| X) from the data maximization joint probability $P(X, Y)$, as the prediction model[42]. That is, the data possibility model built on the global state $P(Y|X) = P(X, Y) / P(X)$. The generative method tries to find out how the data is generated. Generally, it can learn a model representing the target, and search the image region through the target, then classify a signal and minimize the reconstruction error. Based on this generation model, finding the target which is similar to the description of the generated model, and then make template matching to find the most matching region in the image, that is the target in current frame. The specific steps are shown in figure 3 [43].

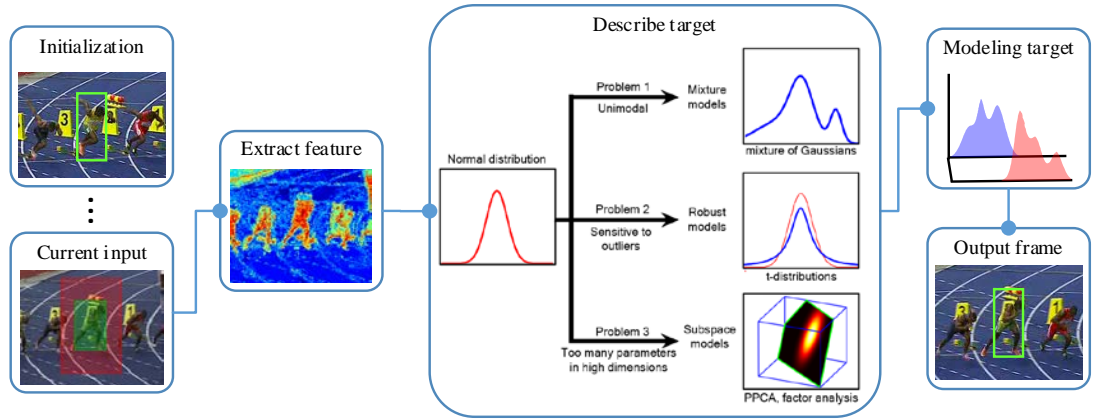

**Fig. 3. Generative method tracker framework.** First, input the video frame and select the target to initialize it. In addition, extracting the target features in the current frame. Then the model is described according to the features of the target and establishing the probability density distribution function. What's more, searching for the next frame of the image region and making template matching to find the region with the highest similarity to the model in the image. Finally, output the target bounding box.

In the framework of visual tracker, the step of extracting target features in the process of target description is very important, which has great influence on the accuracy and speed of tracking. It is not only the generative method applied to feature extraction, but also one of the important steps of model checking in discriminative method. See Table 2 for commonly used feature representation.

As shown in figure 3, describing and modeling target are important steps in the generative method, which can affect the efficiency and accuracy of tracker. Depending on the degree of difficulty in target, the ways of model describe methods are different. The commonly used describe methods include kernel trick [44, 45], incremental learning [46], Gaussian mixed model [47], linear subspace [48], Bayesian network [49], sparse representation [50], hidden Markov model [51] and so on. Finally, the similarity measure function is used as the confidence index to reflect the reliability of each tracking result to determine whether the target is lost or not.

## 2.3 Discriminative Method

The basic idea of discriminative method is that using the data direct learning decision function $Y = f(X)$ or maximization conditional probability distribution $P(Y|X)$ as the prediction model in the learning process. The step is to establish the discriminant function (posteriori probability function) under the condition of finite sample, and to establish the possibility model of data $P(Y|X)$ in the global state, without considering the generation model of the sample, but studying the prediction model directly[42]. In computer vision, this method usually uses the idea of image feature with machine learning. After extracting the target feature, the classifier is trained by the machine learning method to distinguish the target from the background. The architecture of the discriminant class tracking method is shown in figure 4. Because background information is added to the training, the background and target can be distinguished significantly, the performance is more robust, and gradually occupies the mainstream position in the field of visual tracking.

In computer vision, target tracking and target detection are two important parts. The purpose of detection is to find the static or dynamic target in video, and tracking is to locate the dynamic target. The tracking algorithm was originally used to solve the speed of the detection algorithm. It was used to predict the location of the target in the next frame, then the detection algorithm was used to mark the location of the target. Later, some people segment the video sequence according to a certain period of time, and detect each frame image in this period, so the detection can achieve the effect of similar tracking. Such tracking is equivalent to detecting each frame, which is a kind of pseudo-tracking. Tracking developed into "dynamic detection", also known as **Tracking-by-Detection**, which is the mainstream research direction of visual tracking nowadays[52].

Table 2. Recent advances on visual descriptors.

| Feature | Descriptor | Advantage | Representative Method |
|---|---|---|---|
| Grayscale Feature (Histogram) | IH[53], HOI [54], DFs [55] | Earliest, simplest, most intuitive, very fast. | EDFT[56], CSK[57], Frag [29] |
| Gradient Feature (Histogram) | HOG[58], SIFT[59] | The geometric and optical deformation can be kept invariant. | DSST[60-62], CSR-DCF[63] |
| Spatio-Temporal Feature | CMF[64], SOWP[65], CA[66] | Under the fixed background, it has good real-time and robustness for occlusion. | STC[67], CACF[66] |
| Texture Feature | Gabor wave[68], LBP etc. [69-72], WLD[73] | Grayscale invariance and rotation invariance. | TLD[74] |
| Color Feature | CN[75], CBP[76], CC[77] | Strong robustness to photometric changes. | CN[75], DAT[78], ASMS[79] |
| Haar-like Feature | Haar-like[80], Haar-like extra[81] | Very fast and can be calculated in constant time at any scale. | MIL[82], CT[83] |
| Deep Feature | Conv. feat[84] | The best features in state-of-the-art. | CNN-based Tracker |
| Multiple Features Fusion | - | To improve the overall performance or robustness of the system by complementing various methods. | HOG-LBP[85], TOFF[86] (LAB+HOG+LBPF) |

There are usually two kinds of tracking-by-detection methods: one is the **Correlation Filtering** (CF), which trains the filter by regressing the input feature as the target Gauss distribution, and finds the peak value of the response in the prediction distribution to locate the position of the target in subsequent frames [87-91]. The other is the **Deep Learning** (DL), which by updating the weights of the foreground and background in the classifier, it can improve the ability to distinguish the target from its neighborhood background [92-94].

In recent years, a large number of machine learning methods have been modified to deal with the problem of tracking-by-detection, as a training classifier method. In classifier training, supervised learning and semi-supervised learning are commonly used in machine learning, while unsupervised learning is less used (Table 3).

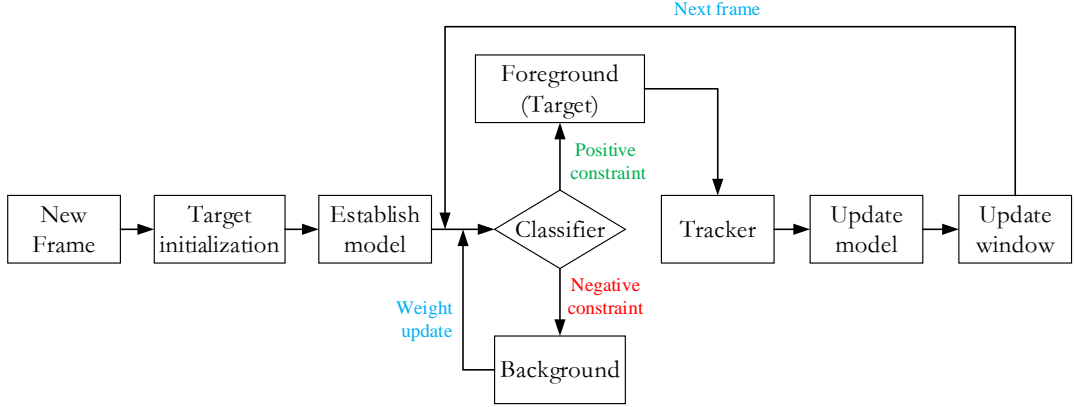

**Fig. 4. Discriminative method framework.** The discriminative method does not care how the data is generated, it only cares about the difference between the signals, it regards the tracking problem as a binary classification problem, and then simply categorize a given signal by difference. Generally speaking, it is the decision boundary to find the target and the background. Tracking is regarded as a frame-by-frame detection problem, and the target frame is selected from the first frame manually.

Table 3. Common machine learning methods

| Method | Representative |
|---|---|
| Integrated learning | Stacking [95], Bosting [96], Adaboost [97], Random Forest [98] |
| Online learning | Co-Training [99], Multi-Instance Learning [82], SVM [100], KNN [101], EM [102], P-N Learning [103] |
| Random learning | CRF [104] |
| Deep learning | CNN [105], DBN [106], SAE [107], RBN [108], R-CNN [109] |
| Bayes classifier | Naive Bayes [110], TAN [111], BAN [111], GBN [112] |
| Regression network | Linear Regression [113], Logistic Regression [114] |

# 3. The Development of Visual tracking

## 3.1 Correlation Filter

Correlation filter (CF), also called discriminative correlation filter (DCF), the principle is that the convolution response of two correlated signals $f$ and $g$ is greater than that of uncorrelated signals (1). Where $f^*$ is the complex conjugate of $f$, the $\int$ is used in continuous domain and the $\sum$ is used in discrete domain. In visual tracking, the filter only generates a high response to each object of interest and a low response to the background. Due to the introduction of circulant matrix and the application of Fast Fourier Transform (FFT), Discrete Fourier Transform (DFT) and Inverse FFT (IFFT), the speed of visual tracking is greatly improved. Computational complexity dropped from $\mathcal{O}(N^2)$ to $\mathcal{O}(N \log N)$.

$$(f \otimes g)(\tau) = \int_{-\infty}^{\infty} f^*(t)g(t+\tau)dt \quad (f \otimes g)[n] = \sum_{-\infty}^{\infty} f^*[m]g[m+n] \tag{1}$$

Since Bolme et al. learn average of synthetic exact filters (ASEF) [115] and minimum output sum of squared error (MOSSE) filter [116], Correlation Filter-based Trackers (CFTs) have attracted considerable attention in the visual tracking community [57, 117] in the following years. Chen et al. [118] summarized the general framework for correlation filtering visual tracking methods in recent years (Figure 5). Most of the current CFTs are based on this framework, and only improve or replace one part of this without affecting the structure of the entire framework. MOSSE only use single channel gray features and shows the high-speed of 615FPS, which shows the advantage of correlation filtering. Then CSK [57] extended the padding and circulant matrix based on MOSSE. After Galoogahi et al. learn MCCF [119] with multi-channel feature, the improved multichannel feature version Kernel Correlation

Filter (KCF) [117] by CSK whose Precision and FPS outperform the best (Struck [120]) on OTB50 [7] at that time (Table 4). CN [75] extends the color feature Color Names based on CSK. With the increase of feature channels, from MOSSE (615FPS) to CSK (292FPS), KCF (172FPS), and CN (152FPS), the speed of tracker is decreasing gradually, but the effect is getting better and better, and it can always be kept at the real-time high speed level. CSK [57], KCF/DCF [117] and CN [75], which have been used as the benchmark in various databases, are correlation filter-based trackers. In the VOT2014 visual tracking competition, the correlation filter-based tracker [62, 117, 121] occupies the top three. Since CSK is learned, the sparse representation-based trackers [83, 122, 123] have gradually been replaced by faster and simpler CFTs.

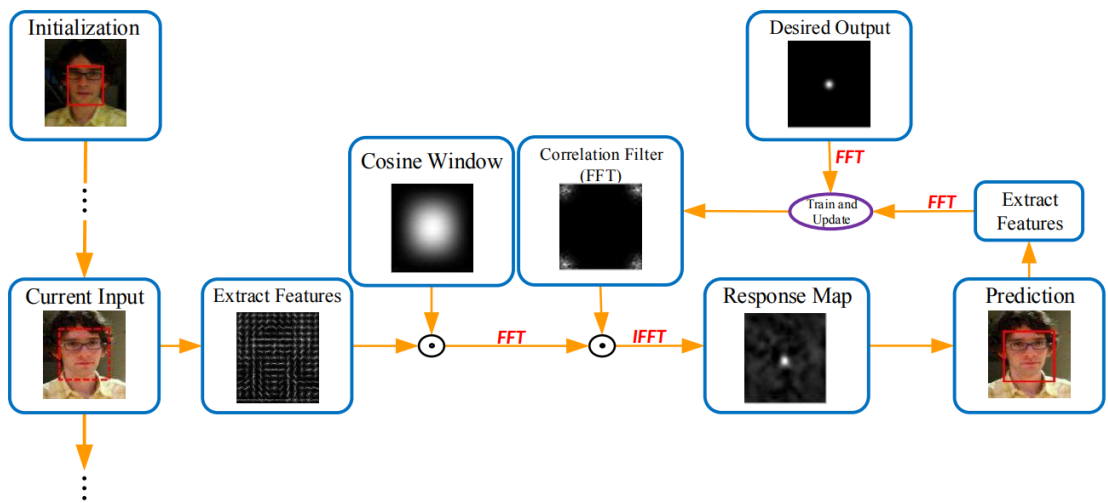

**Fig. 5. General framework for correlation filter visual tracking methods.** After the first frame initialization, in each subsequent frame, an image patch at previously estimated position is cropped as current input. Subsequently, the input can be described better by extracting different visual features and cosine window is usually used to smooth the boundary effect of the window. Afterwards, convolution theorem, the correlation between input signal and the learned filter is obtained by Convolution Theorem. FFT is used to convert the signal into the frequency domain, and the symbol ⊙ in the figure denotes element-wise computation. After the correlation, a spatial confidence map is obtained by IFFT, whose peak can be predicted as the new position of target. Lastly, the feature of the new estimated position is extracted to train and update the correlation filter with a desired output.

Table 4. CSK-based compared to the state-of-the-art tracker at that time

|  | CSK-based | | | | Other | | |
| --- | --- | --- | --- | --- | --- | --- | --- |
|  | KCF [117] | | DCF [117] | | Struck [120] | TLD [74] | MOSSE [116] |
| Tracking Feature | HOG | Raw pixels | HOG | Raw pixels | | | |
| Average accuracy | 73.2% | 56.0% | 72.8% | 45.1% | 65.6% | 60.8% | 43.1% |
| Average FPS | 172 | 154 | 292 | 278 | 20 | 28 | 615 |

Using better feature layers will cause the tracker slow down, and the filter size is fixed, which makes it impossible to respond well to the scale change of the target. So many researchers are focusing on improving the relevant filtering framework. Danelljan et al. proposed DSST [62] with only HOG features, and created a filter architecture based on translation filter combine with scale filter. DCF is used as the filter to detect the translation and the correlation filter similar to MOSSE is trained to detect the scale change of the target. However, the regression formula of DSST is a local optimal problem because the translation filter and the scale filter are solved separately, so that its real-time performance is not good (25FPS). To overcome this problem, Danelljan et al. proposed an accelerated version of f-DSST [61] using PCA dimensionality reduction, which reduces 33 scales to 17, and improves running speed (54FPS). Yang Li et al. proposed SAMF [121] based on KCF which similar to DSST and used HOG add CN features. The image patch is zoomed at multiple scales and then the target is detected by a translational filter. Different from DSST, SAMF combines scale estimation with position estimation to achieve global optimization by iterative optimization. Kiani et al. proposed a type of tracker based on MOSSE, by adding mask matrix P, the filter can crop the real small size samples from large circular shifted patches, so as to increase the proportion of the real sample, which includes CFLB [124] based on grayscale feature and BACF [125] based on HOG feature. Both of them can run in real time (CFLB-87FPS, BACF-35FPS). Sui et al. proposed RCF [126] used three sparse correlation loss functions in the original structure of CF, which can improve the robustness of tracking and real-time performance well (37FPS). Zhang et al. found new ways of using trackers, they proposed MEEM [127] which is essentially a combined tracker. It can call multiple trackers at the same time, and select the best trackers according to the calculation of cumulative loss function, but the actual operation effect is general (13FPS).

The CF template matching method has poor tracking effect on fast deformation and fast motion of target, color feature is not good for illumination change and background similarity, and their performance is unsatisfactory when they are used alone. Bertinetto et al. learned Staple [128] combines template based feature method DSST and color histogram feature based method DAT [78] (15FPS). They found that the accuracy and speed of the tracker combined with the advantages of strong robustness of HOG features to light variation and insensitivity of CN features to deformation were higher than those of the single two trackers. The combined tracker speeds up to 80FPS. Since then, HOG and Color Names have become the standard of Hand-Crafted features in tracking algorithm. Then Bertinetto et al. proposed Staple+ [128] to improve the tracking performance, it increases the number of feature channels from 28 to 56, and adds the response terms of large displacement optical flow motion estimation to the translation detection. Performance has improved, but at the cost of not being real-time. In the same way, Lukezic et al. proposed CSR-DCF [63], combined with the ideas of DAT and CFLB. Using the mask matrix P of CFLB and adding adaptive coefficient, then the response point is determined by CF response map and color probability weighted summation. The maximum response point is determined by weighted sum of CF response map and color histogram. The effect is impressive but the speed is only 13FPS.

Boundary effect has always been one of the difficulties in visual tracking, because of the fast motion, the real samples will escape from the cosine window, so the background will be trained to the classifier, resulting in the sample being contaminated and the tracking failure. In order to solve this problem, Danelljan et al. proposed SRDCF [129], learning the spatial regularization term to punish the filter coefficients in the boundary region and suppressed boundary effect. However, the optimization iteration without closed solution causes the tracker cannot achieve real-time (5FPS). Gundogdu et al analyzed the disadvantages of cosine window and proposed a new window function SWCF [130], which can suppress the irrelevant region of the target and highlight the part of the relative region of the target. However, due to the complexity of the new window function, the speed of the tracker is only 5 FPS. Hu et al. proposed MRCT [131], a manifold regularization-based correlation filter. A regression model is established by using augmented samples and unsupervised learning training classifiers, and similar to BACF, augmented samples are generated from one positive sample cropped in the target region and multiple negative samples cropped in the non-target region, which aims to reduce boundary effect. Bibi et al. proposed CF+AT [132] framework, the target response can be regularized by replacing the samples generated by cyclic shift measurement through actual translation measurement, so as to solve boundary effect. Mueller et al. proposed a Context-Aware based correlation-filter framework CACF [66], which can be used in the learning phase of traditional CF, and the framework can be widely used in many different types of CFTs. CF+AT and CACF improved the performance of tracker significantly, but the speed of tracker is also affected by the increase of computing time.

Tracking confidence is one of the necessary parts of the tracker, which is used to judge whether the target is lost or not. The generative method usually uses similarity measure function, and discriminative method has the classification probability provided by classifier trained by machine learning method. In general, CFTs always use the Maximum Response Peak (MRP, 2, per-channel) $R_{max}$ as the confidence parameter, but it is difficult to effectively determine the target location in complex environment. The earliest correlation filtering method (MOSSE) used Peak to Sidelobe Ratio (PSR, 3) combined with MRP to judge confidence level. Wang et al. proposed LMCF [133] (85FPS) is based on the hand-crafted features and Deep-LMCF (8FPS) based on CNN features. It combined that structure SVM with CF, and proposed Average Peak-to-Correlation Energy (APCE, 4), which can effectively deal with the target occlusion and loss. Yao Sui et al. proposed PSCF [134] based on RCF [126], used a new metric method to enhance the Peak-Strengthened (PS, 5), which is used to improve the discriminative ability of the correlation filters. The tracker can run at 13PFS on desktop. Lukezic et al. believe that the detection reliability of per-channel is reflected in the performance of the major mode value in the response of each channel, so they put forward the Spatio Reliability (6) in CSR-DCF [63]. By combining with the MRP, this tracker performed 13FPS.

$$R_{max} = \zeta \max(\mathbf{f}_d * \mathbf{S}_d) \quad (2)$$

where $\mathbf{f}_d$ donates a filter, $\mathbf{S}_d$ donates discriminative feature channel, the normalization scalar $\zeta$ ensures that $\sum_d R_d = 1$.

$$PSR = \frac{g_{max} - \mu_{sl}}{\sigma_{sl}} \quad (3)$$

where $g_{max}$ is the peak values and $\mu_{sl}$ and $\sigma_{sl}$ are the mean and standard deviation of the sidelobe.

$$APCE = \frac{|R_{max} - R_{min}|^2}{mean\left(\sum_{w,h}(R_{w,h} - R_{min})^2\right)} \quad (4)$$

where $R_{max}$, $R_{min}$ and $R_{w,h}$ denote the maximum, minimum and the w-th row h-th column elements of the peak value of the response.

$$PS = \frac{1}{n}\left(\sum_{j=1}^{n}(R - R_j)^2\right)^{\frac{1}{2}} - \left\|\begin{bmatrix}x_p\\y_p\end{bmatrix} - \begin{bmatrix}x_{gt}\\y_{gt}\end{bmatrix}\right\|_2 \quad (5)$$

where $R$ denotes the peak value of the response, $R_j$ denotes the $j$th response value, n denotes the number of the neighboring response values around the peak, and $[x_p, y_p]^T$ and $[x_{gt}, y_{gt}]^T$ denote the positions of the response peak (correlation output) and the ground truth peak (center of the target location), respectively.

$$R_d^{(det)} = 1 - \min\left(\frac{R_{max2}}{R_{max1}}, \frac{1}{2}\right) \quad (6)$$

Spatio Reliability is based on the ratio between the second and first major mode in the response map. And the per-channel detection reliability is estimated as (6).

Most of the CFTs only pay attention to the performance of short-term tracking, but do not consider long-term tracking that the target will occlude or disappear at any time. Kalal et al. first proposed a novel long-term tracking framework TLD (Tracking-Learning-Detection) [74], which adopts Median-Flow tracker for tracking, P-N learning mechanism and the random fern classifier for detection. Although TLD does not use CF, it provides the original idea for long-term tracking, and the tracker can run in real time. Ma et al. proposed LCT [135], based on the translation filter and scale filter of DSST, added a third correlation filter responsible for detecting the target confidence. It adopted random fern classifier in TLD as the online detector, the running speed is 27FPS. Ma et al. further proposed LCT+, a filter with long-term and short-term memory, added Online SVM Detector and CNN features. LCT+ based on hand-crafted features operating at 20 FPS and 14 FPS by using CNN feature. Hong et al. proposed MUSTer [136] with long-term and short-term memory based on Atkinson-Shiffrin memory model, performed well but runs very slowly (0.287FPS). Zhu et al. proposed a novel collaborative correlation tracker (CCT) [137] using Multi-scale Kernelized the Correlation Tracking (MKC) and Online CUR Filter for long-term tracking. Through the detection of the CUR[1] filter, the drift problem

---

[1] CUR approximation of a matrix A consists of three matrices, C, U, and R, where C is made from columns of A, R is

caused by the long-term occlusion or disengagement of the model is reduced. And the tracker can reach 52FPS.

As can be seen from the above work, the main research direction of CFTs is as follows: (1) Adopt better learning methods; (2) Optimize the regression equation; (3) Extract more powerful features; (4) Reduce the impact of scale change; (5) Weaken the impact of boundary effects; (6) Use better confidence criterion; (7) Combined with the long-term target memory model, etc.

## 3.2 Deep Learning

In recent years, Deep Learning (DL) has been widely concerned [84]. As a representative algorithm, CNN has achieved amazing results in image and speech recognition with its powerful feature expression ability after a series of development [108, 138-141]. In the field of visual tracking, most of DL-based trackers belong to discriminative method. Since 2015, from the top international conferences (ICCV, CVPR, ECCV), it can be seen that more and more DL-based trackers have achieved surprising performance [11].

CNN-SVM [142], proposed by Korean POSTECH team, is one of the earliest DL-based tracker, which combined Convolution Neural Network (CNN) with Support Vector Machine (SVM) classifiers. Finally, the target-specific saliency map is taken as the observation object, tracking is performed by sequential Bayesian filtering. After that, a large number of CNN-based trackers (CNTs) have sprung up. MDNet [143] as an improvement of CNN-SVM, extracted the features of motion with deep learning and added motion features to tracking process. It shows people the potential of CNN in the field of visual tracking, but the tracker is only suitable for running on desktop computer or server, not for running on ARM. In order to improve speed of DL-based method, Held et al. proposed the first DL-based tracker can run at 100FPS[2]. In order to improve the speed, it takes advantage of the large amount of data offline training and avoids online fine-turning, then it doesn't classify patch in regression-based approach, but rather regresses the bounding-box of object. However, these measures can obtain higher FPS, but the price is lower tracking accuracy.

Bertinetto et al. proposed SiameseFC (SiamFC) [144] using Siamese architectures (Figure 6). It is the first tracker to train samples with VID [4] dataset. It performs better than GOTURN and SRDCF in that time, and runs at very fast speed on GPU (SiamFC 58FPS and SiamFC-3s 86FPS). On VOT2016, ResNet-based SiamFC-R and AlexNet-based SiamFC-A outperform, and it is the winner of speed testing on VOT2017 [9, 10]. SiamFC has been attracted a lot of attention because of its excellent performance. It can be said to have opened up another direction for DL-based visual tracking, and the VID dataset also becomes the standard training database of DL-based trackers due to it very suitable for pre-training. In just one year there are such good work to follow up [145-150]. From the results of VOT2017[9], it can be seen that the SiamFC series is a few surviving End-to-End offline training tracker, which is the only direction that can counteract CFTs at present, and it is the most promising direction that can benefit from big data and DL.

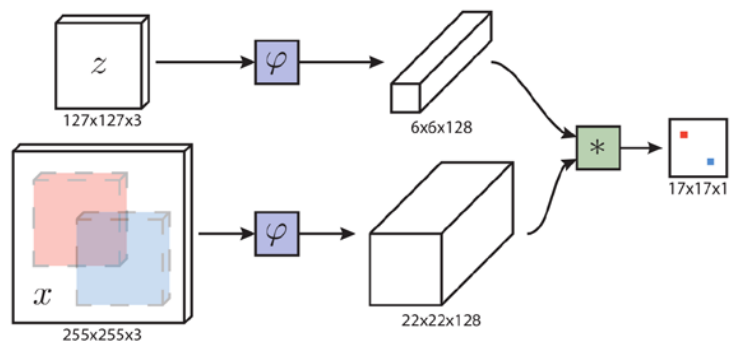

**Fig. 6. Fully-convolutional Siamese architecture.** SiamFC learns a function $f(z, x)$ that compares an

---

made from rows of A, and that the product CUR closely approximates A.
[2] All DL-based trackers use GPU speed.

exemplar image *z* to a candidate image *x* of the same size and returns a high score if the two images depict the same object and a low score otherwise. $\varphi$ is fully-convolutional with respect to the exemplar and candidate image. The output is a scalar-valued score map whose dimension depends on the size of the candidate image. Then computing the similarity responses of all translated sub-windows within the search image in one evaluation, and learn a metric function *g* according to $f(z,x) = g(\varphi(z), \varphi(x))$. Finally, the target position is determined by metric function *g*.

Due to the structural property of CNN, its running speed is always limited. After that, many researchers have proposed combining CF with CNN to speed up the tracker. Bertinetto et al. proposed an improved work CFNet [145] for SiamFC, in this work, they deduced the differentiable closed solution of CF, so that it becomes a layer of CNN. CF is used to build the template of the filter in SiamFC. Then CNN-CF can be used for End-to-End training, which is more suitable for the convolutional features of CF tracking. Tracker can run 43FPS when used conv5. Meanwhile, Wang et al proposed DCFNet [146], used CNN feature instead of HOG feature in discriminative correlation filters (DCF). Besides CNN feature, the other parts are still fast calculated in the frequency domain. The feature resolution is nearly 3 times higher than that of CFNet, and the positioning accuracy is higher. The speed of tracker is 60FPS, but the boundary effect limits the detection area. The latest version of DCFNet 2.0, which has been trained with VID, has made a significant leap forward in performance over CFNet, and operating at 100FPS on GPU. CFCF [151] (the winner of VOT2017 challenge), proposed by Gundogdu et al., had also constructed CNN, that can be End-to-End training based on VID dataset. Unlike the previous trackers, CFCF used the CNN of this fine-tune to extract convolutional features, the rest is exactly the same as C-COT, and this tracker cannot be real-time. Fan et al. proposed PTAV [152], used SiamFC combined with f-DSST, multithreading technology, and drew on the experience of parallel tracking and mapping in VSLAM, uses a tracker T and a verifier V to work in parallel on two separate threads. Through validator to correct the tracker, this problem is studied from a new point of view, and a good experimental result (25FPS) is obtained. There are also a lot of many studies done by Korean Perception and Intelligence Lab on CNN-CF method [153-156], which used Random Forests, Deep Reinforcement Learning, Markov Chains and other machine learning algorithms to optimize the accuracy of classifier, but both of them cannot reach real time.

Huang et al. proposed the first CPU-friendly CNTs EArly-Stopping Tracker (EAST) [147], also an improvement on SiamFC. It tracks simple frames (similar or static) with simple features (HC), while complex frames (obvious changes) use stronger convolutional features to track. The advantage of this is that the average speed of the tracker reaches 23FPS, where 50 % of the time can operate at 190 FPS. On the other hand, the complex frame tracking that needs for convolutional features is very slow, which also shows that the frame rate fluctuation of the tracker will be large. Tao et al. proposed SINT [157] based on Content Based Image Retrieval (CBIR), which only uses the original observation of the target from the first frame. The matching function is obtained by offline training, and Siamese network is used to track the patch which is the best match to the target of initial frame calibration according to the matching function. In the experiment, SINT added optical flow tracking module (SINT+), the effect was improved, but neither of them could run in real time. Wang et al proposed SINT++ [158], which adds positive sample generation network (PSGN) and hard positive transformation network (HPTN) to improve the accuracy of the samples. Although the method is novel and it used the most popular Generative Adversarial Networks (GAN), the actual effect is not impressive.

Chen et al. put forward CRT [159] is different from the traditional DCF in that it does not need to obtain the analytical solution of the regression problem. It attempts to obtain an approximate solution by gradient descent method and a single convolutional layer to solve regression equation. Since convolution regression is trained only on "real" samples without background, it is theoretically possible to incorporate unlimited negative samples. The UCT [150] proposed by Zhu et al. regarded the feature extraction and tracking procedure as a convolution operation, so as to form a completely convoluted network architecture. Similarly, using stochastic gradient descent (SGD) to solve the ridge regression problem in DCF, and using offline training of CNN to accelerate. Meanwhile, they learned a new confidence parameter Peak-versus-Noise Ratio (PNR, 7), and proposed standard UCT (with ResNet-101) and UCT-Lite (with ZF-Net) can operate at 41FPS and 154FPS. Song et al proposed CREST [160], which also reformulated DCF as a one-layer CNN, and uses neural networks to integrate End-to-End training on feature extraction, response graph generation and model update. They learned that features are transformed into the response map through the base and residual mappings for better tracking performance. Park et al. proposed Meta-Tracker [161], an offline meta-learning-based method to adjust the initial deep networks used in online adaptation-based tracking. They demonstrated this approach on CNN-based MDNet [143] and CNN-CF-based CREST [160], then model training speed improved significantly. Yao et al. investigated the joint learning of deep representation and model adaptation on the basis of BACF [125], then proposed RTINet [162], which can run at 9FPS and get a real-time speed of 24 FPS in rapid version.

$$PNR = \frac{R_{max} - R_{min}}{mean(R_{w,h}/R_{max})} \qquad (7)$$

Table 5 collates the network evaluation database maintained by Wang[3] et al. showing the top 20 best-performing tracker at this stage, including CVPR2018. Except for the CF-based tracker BACF and the HC-based tracker ECO-HC (Turbo BACF speed can be over 300FPS, but the source code is not open[4]), the rest of the trackers are based on DL framework, and most of them are based on CNN, but frame rate is generally in single digits. PTAV (SLAM-based), SiamRPN (Siamese network-based) and RASNet can achieve real-time (GPU speed).

Table 5. The trackers are ordered by the average overlap scores.

| Tracker | AUC-CVPR2013 | Precision-CVPR2013 | AUC-OTB100 | Precision-OTB100 | Deep Learning | Real-Time |
|---|---|---|---|---|---|---|
| MOSSE [116] | – | – | 31.1 | 41.4 | N | Y(615) |
| UPDT [163] | – | – | *71.3* | *93.2* | Y | – |
| ECO [164] | *70.9* | 93.0 | *69.4* | 91.0 | Y | N(8) |
| CFCF [151] | *69.2* | 92.2 | 67.8 | 89.9 | Y | N(1.7) |
| LSART [165] | – | – | 67.2 | *92.3* | Y | N(1) |
| MDNet [143] | *70.8* | *94.8* | 67.8 | 90.9 | Y | N(1) |
| SANet [166] | 68.6 | *95.0* | *69.2* | *92.8* | Y | N(1) |
| BranchOut [155] | – | – | 67.8 | 91.7 | Y | N(1) |
| TCNN [167] | 68.2 | *93.7* | 65.4 | 88.4 | Y | N(1) |
| C-COT [168] | 67.2 | 89.9 | 68.2 | – | Y | N(0.3) |
| TSN [169] | – | – | 64.4 | 86.8 | Y | N(1) |
| RASNet [170] | 67.0 | 89.2 | 64.2 | – | Y | *Y(83)* |
| ECO-HC [164] | 65.2 | 84.7 | 64.3 | 85.6 | *N* | *Y(60)* |
| CRT [159] | – | – | 64.2 | 87.5 | Y | N(1.3) |
| BACF [125] | 67.8 | – | 63.0 | 77.6 | *N* | *Y(35)* |
| MCPF [171] | 67.7 | 91.6 | 62.8 | 87.3 | Y | N(0.5) |
| SiamRPN [172] | – | – | 63.7 | 85.1 | Y | *Y(160)* |
| CREST [160] | 67.3 | 90.8 | 62.3 | 83.7 | Y | N(1) |
| DNT [173] | 66.4 | 90.7 | 62.7 | 85.1 | Y | N(5) |
| PTAV [152] | 66.3 | 89.4 | 63.5 | 84.9 | Y | *Y(25)* |
| ADNet [174] | 65.9 | 90.3 | 64.6 | 88.0 | Y | N(3) |

Note: AUC (the area-under-curve) and Precision are the standard metrics. Real Time - FPS, Speeds from the original paper, not test on the same platform. Red - the best, Green - the second, Blue - the third.

Research in recent years has shown that it has always been a difficult point to make the GPU-based real-time trackers run well on CPU. SiamFC [144] cannot real time on CPU because AlexNet will run the same times as the number of scales, which seriously delays the running speed. The fastest DCFNet [146] uses two-layers CNN instead of HOG and the amount of calculation using conv2 is acceptable, but the process of pre-training and fine-tune will make it weak on CPU. EAST [147] as a CNN-based tracker, in most cases it is tracked in the form of KCF, and only in the difficult scenarios will use the conv5 features. In view of that above, if a CNT would perform on CPU or ARM, three points should be noted: (1) It is necessary to control the number of CNN capacity, convolutional layers are the main part of calculation, which needs careful optimization to ensure the speed of CNN. (2) Target image online un-update (no fine-tune), the target features will be fixed after the CNN offline training, thus avoiding the problem that Stochastic Gradient Descent (SGD) and back propagation are almost impossible to real-time in tracking.

---

[3] https://github.com/foolwood/benchmark_results
[4] http://www.hamedkiani.com/bacf.html

## 3.3 Convolutional Features

CFTs have good speed and precision. CNTs have higher accuracy and can keep high speed on GPU. In order to improve the performance of CFTs, it is necessary to adopt the deep feature. CF End-to-end training can be added to the CNTs. CF and DL are not developed independently, they complement and promote each other. The current development direction of tracker is shown in figure 7.

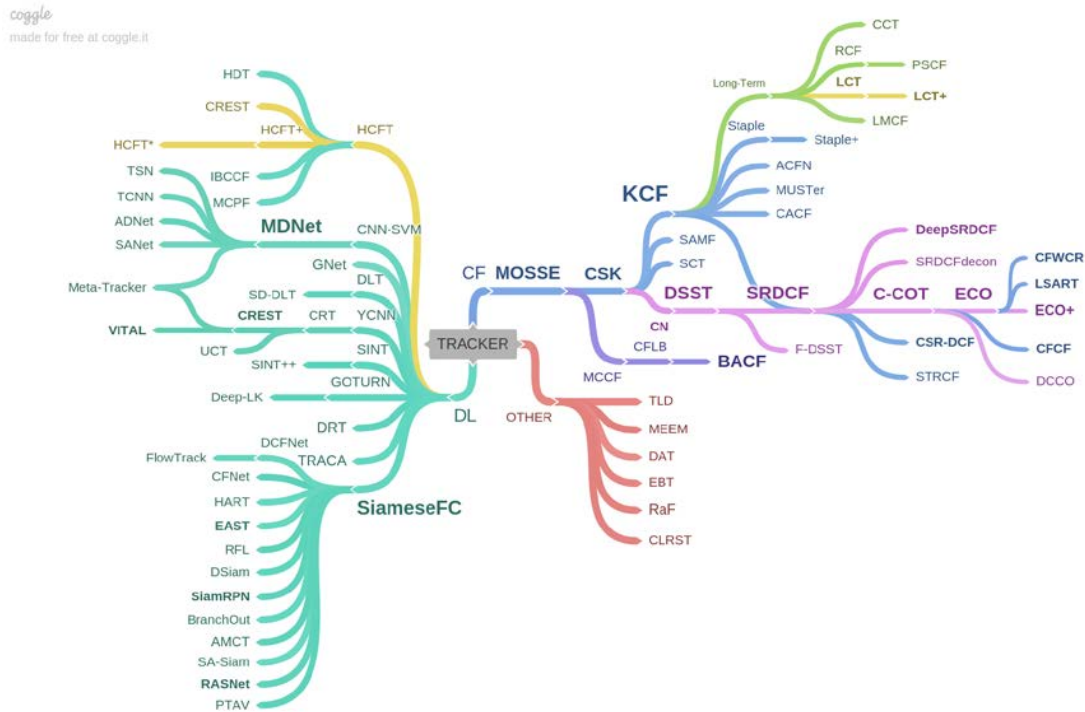

**Fig. 7. The Development Tree of the current trackers.** At present, there are three methods of tracker: (1) CF-based method; (2) CNN-based method; (3) Other. The direction is mainly CF and DL. The **Big Black** fonts represent the stage development. Pink lines represent the contributions of Danelljan et al. Yellow lines represent the contributions of Ma et al.

The right side of the figure 7 represents CF-based trackers. Most of them could be divided into two categories according to the choice of the feature channel. One is to combine the correlation filtering of Hand-Craft features such as HOG, CN or CH (Color Histogram), which can ensure very high speed and good precision, such as BACF [125], ECO-HC [164] and Staple [128]. The others are that CF combined with deep convolutional features, can achieve higher accuracy. Pre-training the convolutional features of CNN model are very strong, generalization ability is very good, but the speed is poor, such as C-COT [168], ECO [164] and CFCF [151].

The left side of the figure 7 represents DL-based trackers, most of them using CNN to train samples, and they can also be divided into two sub categories. Precision oriented MDNet [143] and its extension, can counter the top CFTs on datasets, but due to the limitation of the training set, the generalization ability may be questioned. Speed based SiamFC [144] and its extension, can achieve far more real-time speed on GPU. Especially after the introduction of CF layer, convolutional features extraction can be combined with the detection of CF, and the CNN framework can also achieve intensive detection. Both accuracy and speed can reach a higher level.

A series of work by Danelljan et al. [61, 62, 75, 163, 164, 168, 175] can represent the history of

CFTs, from improving the correlation filtering architecture to solving the boundary effect, to using better features, and then to extracting sub-pixel precision feature. The effect of trackers is getting better and better. They presented a theoretical framework for learning Continuous Convolution Operator Tracker [168] (C-COT), which interpolates feature graphs with different resolution into continuous spatial domain by cubic interpolation. It gets excellent tracking effect, but because of the huge computation, the speed is only 0.3FPS. ECO [164] is an accelerated version of C-COT [168]. It introduced factorized convolution operator, compact generative model and interval update strategy, that simultaneously improves tracking speed and robustness. The GPU version of ECO operates at 8 FPS, and ECO-HC can operate at 60FPS on CPU. On the basis of ECO, He et al. put forward that CFWCR [176] is weighted by double-layer CNN features (conv1 and conv5), and the HC feature is completely abandoned. Although the performance is better than ECO, the cost is to abandon running speed. CFWCR runs at an average of 4FPS on GPU and 1.4FPS on CPU. Bhat et al. analyzed the relationship between the deep and hand-crafted features based on ECO, and proposed UPDT [163], which can make features benefit from the better and deeper CNN layer. It outperforms ECO with a relative gain of 18% on the VOT2016 dataset. Comparing with some state-of-the-art trackers in CVPR2018 [165, 177-179], it still shows the overwhelming advantage. However, UPDT only mentioned the adaptive fusion of feature layer, and did not explain the speed of running. Because of the deeper convolutional features, it should be very slow.

Ma et al. have done a series of works on the use of deep convolutional features. They proposed HCFT [180], with pure convolutional features for tracking, uses the activation values of Conv5-4, Conv4-4 and Conv3-4 in VGG19 as the feature layer and tracks target according to linear weights. It operates at 11FPS on GPU. Then they proposed that HCFT+ [181] and HCFT* [182]. HCFT+ added CF as a part of convolutional layer based on HCFT. By using traditional CF to calculate correlation response diagram on Conv4-4 and Conv5-4 layer, the tracking accuracy is improved and the speed is 12 FPS. HCFT* added a long-term memory filter $w_L$ to HCFT+ for long-term tracking. They proposed a region-based object re-detection and scale estimation scheme. Finally, an incremental updating method for two kinds of CF with different learning rates is proposed. It runs at 6.7FPS and performance better than HCFT+.

In addition to the above two types of CF combined convolutional features tracker, many researchers have proposed more novel methods. Lu et al. proposed LSART [165], the winner in VOT2017, that combines CNN and CF in a new way. They used the iterative method of CF and the regularized kernel in spatial domain to solve CNN, which is more effective than the traditional method. Chen et al proposed a convolutional features-based long-term tracking correlation filter LHCF [183], which is similar to that of HCFT and LCT. The innovation point is to estimate the translation of the target by training the three conventional features layers. Choi et al. proposed TRACA [184], a correlation filter based tracker using context-aware compression of raw deep features. Multiple auto-encoders are used to deal with different category of objects, and the high-dimensional features are compressed into low-dimensional features, which reduced redundancy and sparsity, and improves accuracy and speed. It can run at a fast speed of over 100 FPS.

## 4. Application Analysis of Visual Tracker based on Mobile Robot

From the test results of VOT2017, the high-performance trackers are mainly the following. C-COT [168] used CF combined with conventional features, its accelerated version ECO [164], the fine-tuned version CFCF [151], and the ECO-based GNet with GoogLeNet feature. CPU high speed trackers are ECO-HC [164], Staple [128], ASMS [79], and C++ based CSR-DCF++ [63]. GPU high-speed trackers include SiamFC [144], and its extended version SiamDCF [146], UCT [150]. Although the test results are good, they are all based on test sets. However, for the practical application scenarios, especially the mobile robot, which is the main direction in the future, there are still a lot of difficulties to overcome in the current tracking algorithm.

## (1) Tracking accuracy and speed coordination

What the most important thing for mobile robot is that the visual tracker is more focused on the ability of real-time operation. At the present stage, visual tracking concerns could be divided into two main categories: The first category focuses on improving accuracy, such as MDNet [143], CFCF [151], TCNN [167], etc. This kind of tracker does achieve high precision and high ranking on each data set, but the speed is very slow (both on CPU and GPU), which cannot meet the requirement of mobile robot real-time application. The second focuses on real-time performance, such as Staple [128], ECO-HC [164], EAST [147] and so on, which guarantees accuracy and is much faster than DL-based architecture. From the VOT2017 challenge results, the top ten trackers on public dataset are C-COT-based or ECO-based and the main features used convolutional and hand-crafted features. The performance of the tracker with convolutional features is better than that with hand-crafted features only, but the speed of the tracker is also decreased seriously. Although the performance of the GPU-based trackers is getting better and faster with the great development of deep learning, it has a good performance in the desktop work scene, but it can not be applied to the mobile side (based on ARM or CPU). Whether convolutional features are needed, or whether to find the better-faster features, is what the tracker needs to consider when it comes to mobile robots.

## (2) Combination of target detection and tracking

In the test database of trackers, because all the targets are pre-calibrated, that is, the initial position of the target in each set of video frames is already known before tracking, the tracker tracks calibrated target directly. Therefore, it does not represent the ability to initialize the tracker in practical applications. Human brain has a strong logical reasoning ability, so it can identify the target at any time, and the "first impression" of the target can be quickly stored in the mind, and then can follow it all the time. However, the visual tracker selectively ignores the important issue of how the first frame bounding box comes from. Some trackers often appear to be weak when it is necessary to independently select and track new targets. The next step of visual tracking can be fused with target detection and recognition, which can independently confirm the target and then tracking it. In mobile robot applications, detect to Track would certainly be a closed loop in the future, rather than limited to the performance or speed of the tracker.

## (3) Ability to long-term tracking

Since the 2014 Long-Term Detection and Tracking workshop (LTDT[5]), long-term tracking has been a major concern. A new Long-term tracking sub-challenge[6] has also been added to the VOT2018, which requires the tracker to determine that the target disappears and to re-detect and track it when the target enters the scene again. This shows that the importance of long-term tracking has been paid more and more attention. At present, most visual trackers focus on the accuracy of Short-term tracking (e.g. 100~500 frames). But in practical applications, such as mobile robots, the tracking time is often uncertain, may be a few minutes or a dozen minutes or even longer, a lot of occlusion, target-loss problems are not prominent in short-term, which affects the actual use of the tracker. Therefore, the tracer can be required to Long-term stable tracking. Long-term tracking needs to add re-detector and longtime memory model to the traditional tracker, and they can be called to rectify the tracker if the trace fails. Of course, the short-term tracking performance of tracker is also related to the quality of long-term.

---

[5] http://www.micc.unifi.it/LTDT2014
[6] http://www.votchallenge.net/vot2018/

### (4) Good portability

At present, most of the trackers are based on Tracking-by-Detection. For the performance of visual tracking, the selection of features has a great influence on tracking performance. Danelljan et al. [185] proved that the deep convolution feature has good rotation invariability but the speed advantage will be lost by introducing the convolution feature. Nowadays, the DL-based trackers (including extracted conventional features) take GPU as the core of computing, and need the specialized computing card such as Tesla or Titan to pre-train datasets, which are often composed of multiple graphics cards, which are expensive and power consuming. Table 5 shows that DL-based trackers are also becoming more and more difficult to run on GPU, DL-based trackers cannot benefit from deeper CNN [163]. For the mobile robot, the portability of the controller is an important factor affecting the physical parameters of the mobile robot, such as volume, endurance, structure complexity and so on, so the DL-based tracker is not suitable. Finding the CPU-friendly DL-based tracker (such as EAST [147]) may be a future development direction. Of course, compared with deep learning, CFTs are more suitable for mobile robot at the present stage.

## 5. Conclusion and future directions

Visual tracking as an important component of Computer Vision with many applications which makes it a highly attractive research problem. In this paper, we summarized the difficulties and general architecture of visual tracking. Then we provided a list of visual feature descriptors and summarized machine learning methods about trackers. With the view of real-time performance, state-of-the-art visual trackers based on Tracking-by-Detection were introduced from Correlation Filter, Deep Learning and Convolutional Features-based perspectives. Finally, the key point of application of trackers in mobile robots were analyzed, which is also trackers forthcoming research directions.

Although the generative method framework has the advantages of good real-time performance and less adjustment parameters, its modeling complexity limits its further development. With the development of correlation filter and deep learning, discriminative method algorithm based on Tracking-by-Detection architecture has become the mainstream. Their speed, precision and robustness have completely exceeded the generative method. However, the potential of deep learning in visual tracking direction is not well demonstrated, and replacing different neural networks does not result in substantial performance improvements [163]. Because the architecture determines that computing is inherently slow (on CPU), although the trackers based on DL or based on CF with convolutional features outperforms the CFTs based on HC features by 10% ~ 15%, there is no absolute advantage in practical application, and there is not much gap with CFTs. Instead, the speed of running on the CPU will constrain its performance. All in all, the running speed of computer vision algorithm is one of the most important indexes of algorithm performance, especially the visual tracker, which always puts the speed ahead of the performance in practical application. But in academic research, performance is often emphasized, and real-time testing is neglected. That is to say, the ultimate purpose of the visual tracker should to focus on practical applications, rather than just in their own circle of research to "Benchmark & Tuning".

Unlike fixed position manipulator, the video camera that a mobile robot carries would move along with the robot, and sometimes it will have to rotate itself. So what the tracker needs to locate is the relative position of the target. Similarly, the visual trackers using database to test only collects two-dimensional plane information, which does not collect depth information in space. Depth information is an important physical parameter necessary for mobile robot. So how to convert the vision algorithm suitable for plane tracking into the vision algorithm suitable for space tracking maybe a research direction in the future. For the video camera, the illumination variation is a common problem. The white balance of the video camera will go a sudden change when it is exposed to strong light, which will interfere with the tracking and updating of the target features. The research of high performance visual tracker suitable for mobile robot is not only limited to testing in database, but also needs to combine many kinds of sensors to assist visual tracking, and track target accurately in the open environment that is not restricted by databases and training datasets. Finally, the use of specific environmental information is also an important research direction. Such as vehicle tracking, cars should be kept on the road, not on the sky or on the wall. This kind of semantic or environmental information is also very useful for the development of trackers.

**Acknowledgments:** This research was supported by grant of the National Key Research and Development Program of China (No. 2018YFC0808000) and the Priority Academic Program Development of Jiangsu Higher Education Institutions (PAPD), China.

**Author Contributions:** Shaoze You designed the architecture and finalized the paper. Hua Zhu conceived the idea. Menggang Li and Yutan Li did the proof reading.

**Conflicts of Interest:** The authors declare no conflict of interest.